\documentclass[letterpaper, 10 pt, conference]{ieeeconf}  
\IEEEoverridecommandlockouts                              
\usepackage{epsfig} 
\usepackage{amsmath} 
\usepackage{amssymb}  
\usepackage{color}
\usepackage{cite}
\usepackage{txfonts}
\usepackage{amsmath}
\usepackage{amssymb}
\usepackage[linesnumbered,ruled]{algorithm2e}
\usepackage[normalem]{ulem} 
\makeatletter
\let\NAT@parse\undefined
\makeatother
\usepackage{hyperref}
\hypersetup{pdfstartview=FitH,
            colorlinks=true,
            linkcolor=red,
            citecolor=black
            }
\usepackage{booktabs}
\usepackage{arydshln}
\usepackage{multirow}
\usepackage{mathrsfs}
\usepackage[caption=false,font=footnotesize]{subfig}
\usepackage[dvipsnames]{xcolor}
\usepackage{subcaption}
\usepackage{graphicx}
\usepackage{subfig} 
\usepackage{soul}
\usepackage{ulem}
\newcounter{RNum}
\renewcommand{\theRNum}{\arabic{RNum}}
\newcommand{\Remark}{\noindent\textit{\textbf{Remark}~\refstepcounter{RNum}\textbf{\theRNum}: }}
\newcommand{\NoOne}[1]{\textcolor{red}{#1}}
\newcommand{\NoTwo}[1]{\textcolor{green}{#1}}
\newcommand{\NoThree}[1]{\textcolor{blue}{#1}}
\usepackage{xcolor}
\soulregister\NoOne7
\soulregister\NoTwo7
\soulregister\NoThree7
\soulregister\Remark7

\soulregister{\cite}7 
\soulregister{\citep}7 
\soulregister{\citet}7 
\soulregister{\ref}7
\soulregister{\pageref}7 
\usepackage{amssymb}  
\usepackage{pifont}   
\usepackage[absolute,overlay]{textpos}
\usepackage{multirow}
\usepackage{multicol}
\usepackage{graphicx}
\usepackage{amsmath}
\usepackage{amssymb}
\usepackage{booktabs}
\usepackage{xcolor,colortbl}
\usepackage{makecell}
\usepackage{pifont}
\usepackage{balance}
\usepackage{cancel}
\usepackage{adjustbox}

\definecolor{nbarrier}{RGB}{255, 120, 50}
\definecolor{nbicycle}{RGB}{255, 192, 203}
\definecolor{nbus}{RGB}{255, 255, 0}
\definecolor{ncar}{RGB}{0, 150, 245}
\definecolor{nconstruct}{RGB}{0, 255, 255}
\definecolor{nmotor}{RGB}{200, 180, 0}
\definecolor{npedestrian}{RGB}{255, 0, 0}
\definecolor{ntraffic}{RGB}{255, 240, 150}
\definecolor{ntrailer}{RGB}{135, 60, 0}
\definecolor{ntruck}{RGB}{160, 32, 240}
\definecolor{ndriveable}{RGB}{255, 0, 255}
\definecolor{nother}{RGB}{139, 137, 137}
\definecolor{nsidewalk}{RGB}{75, 0, 75}
\definecolor{nterrain}{RGB}{150, 240, 80}
\definecolor{nmanmade}{RGB}{213, 213, 213}
\definecolor{nvegetation}{RGB}{0, 175, 0}

\definecolor{car}{rgb}{0.39215686, 0.58823529, 0.96078431}
\definecolor{bicycle}{rgb}{0.39215686, 0.90196078, 0.96078431}
\definecolor{motorcycle}{rgb}{0.11764706, 0.23529412, 0.58823529}
\definecolor{truck}{rgb}{0.31372549, 0.11764706, 0.70588235}
\definecolor{other-vehicle}{rgb}{0.39215686, 0.31372549, 0.98039216}
\definecolor{person}{rgb}{1.        , 0.11764706, 0.11764706}
\definecolor{bicyclist}{rgb}{1.        , 0.15686275, 0.78431373}
\definecolor{motorcyclist}{rgb}{0.58823529, 0.11764706, 0.35294118}
\definecolor{road}{rgb}{1.        , 0.        , 1.        }
\definecolor{parking}{rgb}{1.        , 0.58823529, 1.        }
\definecolor{sidewalk}{rgb}{0.29411765, 0.        , 0.29411765}
\definecolor{other-ground}{rgb}{0.68627451, 0.        , 0.29411765}
\definecolor{building}{rgb}{1.        , 0.78431373, 0.        }
\definecolor{fence}{rgb}{1.        , 0.47058824, 0.19607843}
\definecolor{vegetation}{rgb}{0.        , 0.68627451, 0.        }
\definecolor{trunk}{rgb}{0.52941176, 0.23529412, 0.        }
\definecolor{terrain}{rgb}{0.58823529, 0.94117647, 0.31372549}
\definecolor{pole}{rgb}{1.        , 0.94117647, 0.58823529}
\definecolor{traffic-sign}{rgb}{1.        , 0.        , 0.    }    

\makeatletter
\newcommand{\car@semkitfreq}{3.92}
\newcommand{\bicycle@semkitfreq}{0.03}
\newcommand{\motorcycle@semkitfreq}{0.03}
\newcommand{\truck@semkitfreq}{0.16}
\newcommand{\othervehicle@semkitfreq}{0.20}
\newcommand{\person@semkitfreq}{0.07}
\newcommand{\bicyclist@semkitfreq}{0.07}
\newcommand{\motorcyclist@semkitfreq}{0.05}
\newcommand{\road@semkitfreq}{15.30}  %
\newcommand{\parking@semkitfreq}{1.12}
\newcommand{\sidewalk@semkitfreq}{11.13}  %
\newcommand{\otherground@semkitfreq}{0.56}
\newcommand{\building@semkitfreq}{14.1}  %
\newcommand{\fence@semkitfreq}{3.90}
\newcommand{\vegetation@semkitfreq}{39.3}  %
\newcommand{\trunk@semkitfreq}{0.51}
\newcommand{\terrain@semkitfreq}{9.17} %
\newcommand{\pole@semkitfreq}{0.29}
\newcommand{\trafficsign@semkitfreq}{0.08}
\newcommand{\semkitfreq}[1]{{\csname #1@semkitfreq\endcsname}}

\hypersetup{colorlinks = true, 
	    linkcolor = red,
            citecolor = green}

\title{\LARGE \bf
OccRWKV: Rethinking Efficient 3D Semantic Occupancy Prediction with Linear Complexity
}

\author{Junming Wang$^{1,2,*}$, Wei Yin$^{1,*}$, Xiaoxiao Long$^{3,\dag}$, Xingyu Zhang$^{1}$, Zebin Xing$^{1}$, Xiaoyang Guo$^{1}$, Qian Zhang$^{1}$         
\thanks{ $^{*}$Equal Contribution. $^{\dag}$Corresponding Author. }   
\thanks{$^{1}$Horizon Robotics. $^{2}$University of Hong Kong. $^{3}$Nanjing University.} 
}

\begin{document}

\maketitle
\thispagestyle{empty}
\pagestyle{empty}

\begin{abstract}

3D semantic occupancy prediction networks have demonstrated remarkable capabilities in reconstructing the geometric and semantic structure of 3D scenes, providing crucial information for robot navigation and autonomous driving systems. However, due to their large overhead from dense network structure designs, existing networks face challenges balancing accuracy and latency.
In this paper, we introduce OccRWKV, an efficient semantic occupancy network inspired by Receptance Weighted Key Value (RWKV). OccRWKV separates semantics, occupancy prediction, and feature fusion into distinct branches, each incorporating Sem-RWKV and Geo-RWKV blocks. These blocks are designed to capture long-range dependencies, enabling the network to learn domain-specific representation (i.e., semantics and geometry), which enhances prediction accuracy. Leveraging the sparse nature of real-world 3D occupancy, we reduce computational overhead by projecting features into the bird's-eye view (BEV) space and propose a BEV-RWKV block for efficient feature enhancement and fusion. This enables real-time inference at 22.2 FPS without compromising performance. 
Experiments demonstrate that OccRWKV outperforms the state-of-the-art methods on the SemanticKITTI dataset, achieving a mIoU of 25.1 while being 20 times faster than the best baseline, Co-Occ, making it suitable for real-time deployment on robots to enhance autonomous navigation efficiency. Code and video are available on our project page: \url{https://jmwang0117.github.io/OccRWKV/}.

\end{abstract}

\section{INTRODUCTION}

3D semantic occupancy prediction networks\cite{cao2022monoscene,tang2024sparseocc,li2023voxformer} have garnered significant attention in recent years due to their remarkable ability to reconstruct the geometric and semantic structure of 3D scenes, providing comprehensive occupancy maps and semantic information crucial for robot navigation tasks \cite{wang2024agrnav,wang2024he} and autonomous driving systems \cite{huang2023tri,tang2024sparseocc,pan2024co}. 

\begin{figure}[t]
  \centering
     \includegraphics[width=\linewidth]{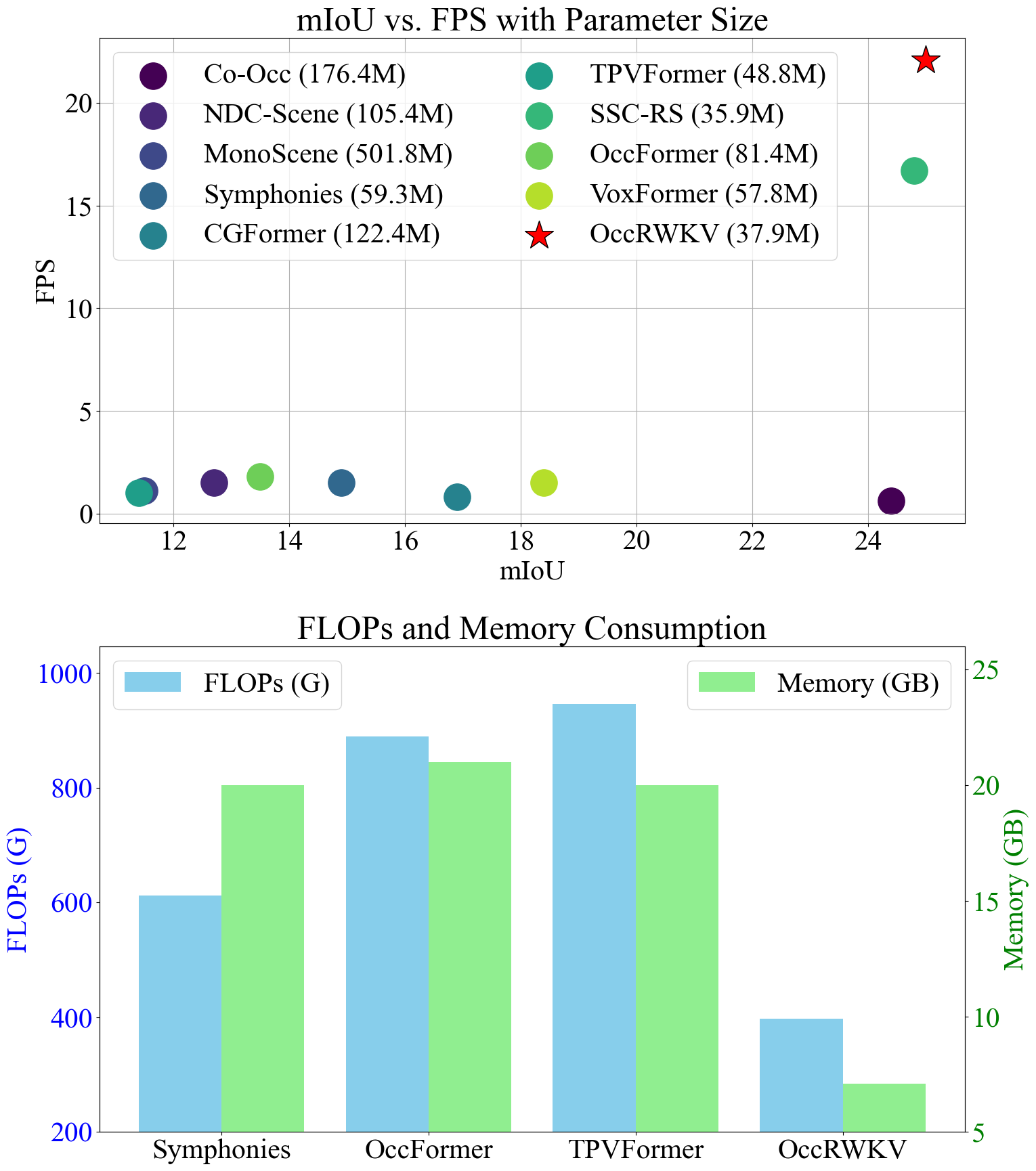}
   \caption{Comparison of accuracy and efficiency metrics (i.e., FPS and FLOPs) with SoTA methods.  }
   \label{fig:overview}
\end{figure}

Although existing \textit{single modality} (i.e., LiDAR-based \cite{wang2024agrnav,roldao2020lmscnet,mei2023ssc,wang2024omega} and Camera-based \cite{cao2022monoscene,li2023voxformer,wei2023surroundocc}) and \textit{multi-modal} networks \cite{pan2024co} have made significant advancements in 3D semantic occupancy predictions, most of them employ dense 3D CNN \cite{cao2022monoscene} or transformer \cite{dosovitskiy2020image} architectures, which have high computational complexity and requires large GPU memories. Such requirements hinder them deployed in resource-constrained environments, such as robotics systems and autonomous driving.

Some methods attempt to reduce network complexity by utilizing 2D convolution \cite{roldao2020lmscnet,wang2024agrnav}. While this approach helps to mitigate the computational burden, it comes at the cost of failing to capture long-range dependencies that are essential for accurate semantic segmentation and occupancy prediction. The inability to effectively model long-range context information limits the performance of these methods, particularly in complex and dynamic environments.

Our key insights to address these challenges lie in rethinking and designing novel network structures that enable 3D semantic occupancy prediction networks to strike a balance between accuracy and latency. Firstly, we recognize that 3D occupancy in the real world is sparse, with most voxels being empty. This sparsity suggests the potential benefits of migrating dense feature fusion to the bird's-eye view (BEV) space \cite{li2023bev,li2022bevformer,liu2023bevfusion,mei2023ssc}, which can lead to more efficient computations and reduced memory requirements. 

Secondly, we draw inspiration from the recent \textit{Receptance Weighted Key Value (RWKV)} model \cite{peng2023rwkv,peng2024eagle}, which utilizes a linear tensor-product attention mechanism. This mechanism avoids quadratic complexity and improves computational efficiency, allowing RWKV to maintain lower memory and computational overhead when processing long sequences. The RWKV model has been successfully adapted for vision tasks in Vision-RWKV (VRWKV) \cite{duan2024vision} by introducing a quad-directional shift (Q-Shift) and modifying the original causal RWKV attention mechanism to a bidirectional global attention mechanism. This adaptation not only inherits the efficiency of RWKV in handling global information and sparse inputs but also models the local concept of vision tasks and reduces spatial aggregation complexity. Inspired by these observations, we pose the following question: \textit{Can we design a 3D semantic occupancy network with linear complexity that achieves a trade-off between performance (i.e., accuracy) and efficiency (i.e., faster inference speeds and lower memory usage)?}

Building upon these insights, we introduce \textbf{OccRWKV}, the first RWKV-based 3D semantic occupancy network. Unlike earlier approaches that combine semantic and occupancy predictions, OccRWKV uses separate pathways for each task. This design allows each branch to focus on its specific learning objectives, improving both semantic and geometric predictions. The system then combines these features effectively in a later fusion step. The architecture includes specialized RWKV blocks for semantics, geometry, and bird's-eye-view processing, which help capture important relationships across different parts of the input. By converting features to a bird's-eye-view format, the system can process data efficiently while maintaining high accuracy.

We first assessed OccRWKV on the SemanticKITTI benchmark, comparing its accuracy and inference speed to some leading occupancy networks. Next, we also deployed OccRWKV on a real robot to test its efficiency in navigation tasks. Our evaluation reveals:

\begin{itemize}
    \item \textbf{OccRWKV is high-performance.} OccRWKV achieves state-of-the-art performance (mIoU = 25.1) on the SemanticKITTI benchmark. (§~\ref{sec:B1})
    
    \item \textbf{OccRWKV is efficient.}  OccRWKV not only runs 20x faster than the best baseline (i.e., Co-Occ), achieving 22.2 FPS with superior performance while reducing the parameter count by 78.5\%. (§~\ref{sec:B1})
    
    \item \textbf{OccRWKV is plug and play.} OccRWKV can be deployed on real robots as an occlusion perception network to improve navigation efficiency. (§~\ref{sec:D1})
    
\end{itemize} 

\begin{figure*}[t]
  \centering
     \includegraphics[width=\linewidth]{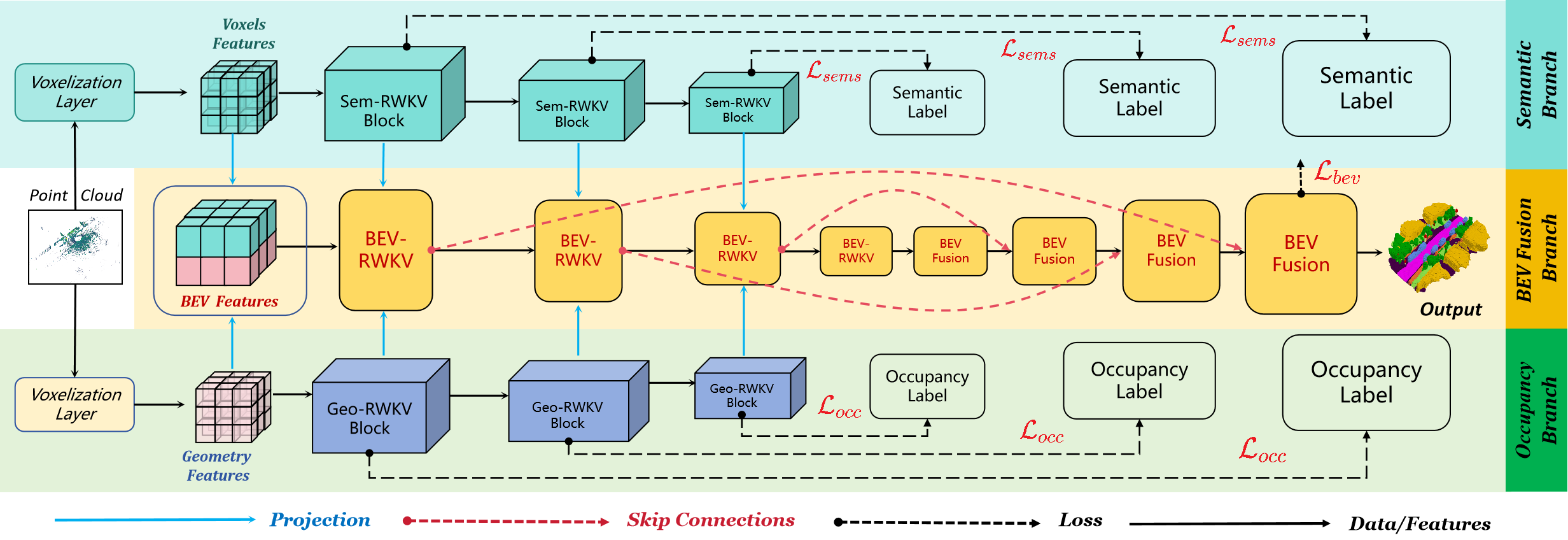}
   \caption{Overview of the proposed OccRWKV. Semantic and geometry branches learn respective representations, i.e., semantic and geometric, supervised by multi-level auxiliary losses. Finally, features are fused in the BEV fusion branch to generate dense 3D semantic occupancy predictions.}
   \label{fig:overview}
\end{figure*}

\section{RELATED WORK}

\subsection{3D Semantic Occupancy Prediction}

3D semantic occupancy prediction \cite{wang2024omega} is crucial for interpreting occluded environments, as it discerns the spatial layout beyond visual obstructions by merging geometry with semantic clues. The field has seen diverse approaches, broadly categorized into \textit{CNN-based} and \textit{Transformer-based} methods. \textit{CNN-based} methods have demonstrated proficiency in inferring occupancy from various inputs. The Co-Occ \cite{pan2024co} framework adopts a multi-modal strategy that fuses LiDAR and camera data, enhanced by volume rendering regularization and a Geometric- and Semantic-aware Fusion module, achieving notable performance on public benchmarks. LowRankOcc \cite{zhao2024lowrankocc} employs tensor decomposition and low-rank recovery to address spatial redundancy, leading to state-of-the-art results on multiple datasets. Other notable works such as JS3C-Net \cite{yan2021sparse} and SSC-RS \cite{mei2023ssc} adeptly manage the complexity of outdoor scenes using point cloud data. \textit{Transformer-based} methods leverage the attention mechanism for feature aggregation and have shown promising results. TPVFormer \cite{huang2023tri}  introduces a tri-perspective approach that combines BEV with two additional planes, achieving LiDAR-like perception using camera inputs alone. 

\subsection{Receptance Weighted Key Value (RWKV) Models}

The Receptance Weighted Key Value (RWKV) model \cite{peng2023rwkv} presents a novel solution to the challenges faced by traditional deep learning architectures in sequence processing tasks. RNNs \cite{yu2019review} struggle with training difficulties for long sequences due to vanishing gradients and limited parallelization. Transformers \cite{vaswani2017attention} have revolutionized the field with their parallel training capabilities and superior handling of dependencies, but their success comes at the cost of high computational and memory demands, especially for longer sequences. RWKV addresses these challenges by integrating the parallel training capabilities of Transformers with the linear computational efficiency of RNNs. It employs a redesigned linear attention mechanism that avoids the costly dot-product interactions of traditional Transformers, enabling efficient channel-directed attention and scalable model performance. This innovative approach allows RWKV to maintain the expressive power of Transformers while providing a more resource-efficient architecture, making it suitable for handling longer sequences without the quadratic scaling limitations.

\subsection{RWKV-Based Approaches in Visual Perception Tasks}

The RWKV model, originally impactful in NLP, has been effectively adapted for visual perception tasks \cite{wang2024neurncd}, highlighting its versatility. Vision-RWKV \cite{duan2024vision} addresses high-resolution image processing with reduced complexity, while PointRWKV \cite{he2024pointrwkv} applies RWKV to point cloud encoding with a hierarchical structure for multi-scale feature capture. Diffusion-RWKV \cite{fei2024diffusion} extends RWKV to image generation, efficiently handling large-scale data and achieving high-quality results with less computational cost. \textit{In this paper, We introduce \textbf{OccRWKV}, the first 3D semantic occupancy network leveraging the RWKV architecture, enabling efficient real-time semantic occupancy prediction and showcasing a novel application of RWKV in 3D spatial analysis.}

\section{Method}

In this section, as depicted in Fig.~\ref{fig:overview}, we dissect the architecture of our proposed \textbf{OccRWKV} into three integral components: the semantic segmentation branch (§~\ref{sec:3a}), the occupancy prediction branch (§~\ref{sec:3b}), and the BEV feature fusion branch (§~\ref{sec:3c}). We culminate the section (§~\ref{sec:3d}) by detailing the training loss function.

\subsection{Semantic Segmentation Branch}
\label{sec:3a}
\noindent\textbf{\textit{Voxelization Layer}:} We partition the 3D environment into voxels for prediction. The semantic component employs a voxelization layer followed by three Sem-RWKV Blocks of identical structure. Our system transforms an input point cloud $P \in \mathbb{R}^{N \times 3}$ within the range $[R_x, R_y, R_z]$ into voxel features $F_V \in \mathbb{R}^{M \times C}$, creating a spatial resolution of $L \times W \times H$. For each point $p_i = (x_i, y_i, z_i)$, we calculate its voxel index $V_i$ \cite{mei2023ssc} using:

\begin{equation}
     V_i = \left( \left\lfloor \frac{x_i}{s} \right\rfloor, \left\lfloor \frac{y_i}{s} \right\rfloor, \left\lfloor \frac{z_i}{s} \right\rfloor \right) 
\end{equation} where $s$ denotes the voxelization resolution and $\left\lfloor \cdot \right\rfloor$ represents the floor function. Considering that multiple points may occupy a single voxel, the voxel features $f_{V_m}$ indexed by $V_m \in \mathbb{Z}^{L \times W \times H}$ are aggregated using:
\begin{equation}
   f_{V_m} = R_f \left( A_f \left( \text{MLP}(f_p)_{V_p=V_m} \right) \right) 
\end{equation} Here, $A_f$ is the aggregation function (e.g., max function), and $R_f$ denotes MLPs for dimension reduction. We construct the point features $f_p$ by concatenating the point coordinates, the distance offset from the voxel center where the point is located, and the reflection intensity.

\begin{figure}[t]
  \centering
     \includegraphics[width=\linewidth]{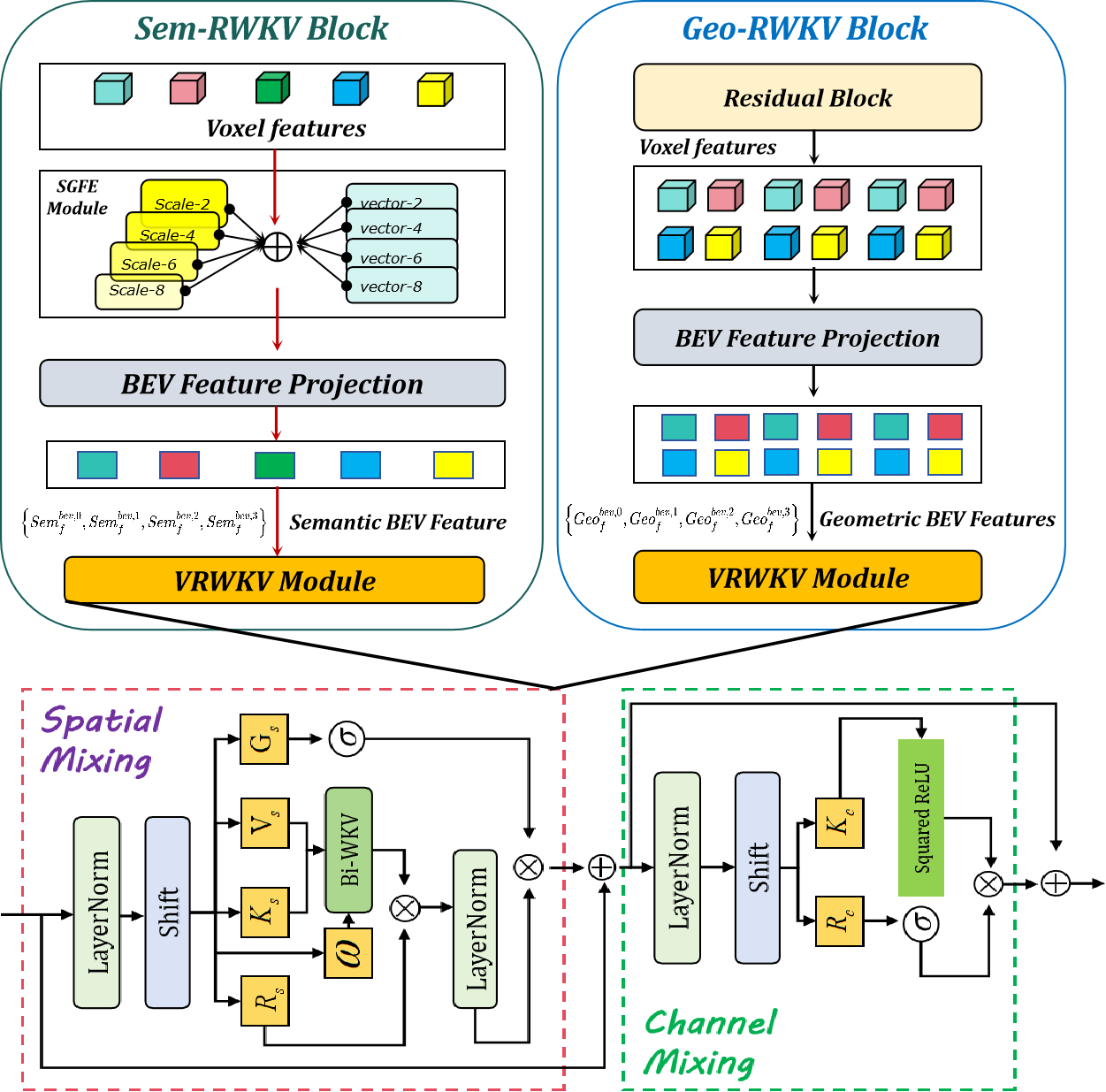}
   \caption{The overview of the proposed Sem-RWKV and Geo-RWKV blocks is illustrated, please zoom in for details. }
   \label{fig:block1}
\end{figure}

\noindent\textbf{\textit{Sem-RWKV Blocks}:} After obtaining the voxel features, we fed them into three cascades of Sem-RWKV encoder blocks (in Fig.~\ref{fig:block1}) to obtain dense \underline{\textit{Semantic-BEV features}}. Each Sem-RWKV block comprises several key components: residual blocks, the sparse global feature enhancement (SGFE) module \cite{xu2022sparse,mei2023ssc} for enriching voxel features with geometric context, a BEV projection module, and a VRWKV module \cite{duan2024vision} for feature enhancement. The SGFE module employs multi-scale sparse projections alongside attentive scale selection, augmenting the geometric details at the voxel level while halving the resolution of dense features, a crucial step for semantic feature extraction. The resulting semantic features $\{Sem_{f}^{1}, Sem_{f}^{2}, Sem_{f}^{3}\}$ are mapped into bird's-eye view (BEV) coordinates, where each voxel is assigned a unique BEV index based on its $f_m$ value. Features with identical BEV indices are then aggregated via max pooling, yielding a collection of sparse BEV features. These sparse features are subsequently densified using Spconv's densification function, producing dense \underline{\textit{Semantic-BEV features}} $\left \{Sem^{bev,0}_f, Sem^{bev,1}_f, Sem^{bev,2}_f, Sem^{bev,3}_f  \right \} $.

The dense \underline{\textit{Semantic-BEV features}} are then processed by the Vision-RWKV (VRWKV) module \cite{duan2024vision}, which comprises two key components: the Spatial Mixing module and the Channel Mixing module. In the Spatial Mixing module, the input features undergo a shifting operation denoted as $Q$-$Shift$, and are projected into matrices $R_s, K_s, V_s \in \mathbb{R}^{T\times C}$ through parallel linear transformations:

\begin{align}
    R_s &= Q\text{-}Shift_R(X)W_R \\
    K_s &= Q\text{-}Shift_K(X)W_K \\
    V_s &= Q\text{-}Shift_V(X)W_V
\end{align}

The global attention output $wkv$ is computed via a linear-complexity bidirectional attention mechanism $Bi$-$WKV$ from \cite{duan2024vision}, applied to $K_s$ and $V_s$:

\begin{align}
    wkv = Bi\text{-}WKV(K_s, V_s).
\end{align} where attention calculation result for the t-th feature token is given by the following formula:
\begin{equation}
\begin{split}
wkv_t &= Bi-WKV(K,V)_t\\
      &\quad\cdot\frac{{\textstyle \sum_{i=0,i\ne t}^{T-1}}e^{-(|t-i|-1/)T\cdot \omega +k_i} v_i+e^{u+k_t}v_t}
                      {{\textstyle \sum_{i=0,i\ne t}^{T-1}}e^{-(|t-i|-1/)T\cdot \omega +k_i}+e^{u+k_t}}
\end{split}
\end{equation}

The output $O_s$ is obtained by element-wise multiplication of $\sigma(R_s)$ and $wkv$, followed by a linear projection and layer normalization:

\begin{align}
    O_s = (\sigma(R_s) \odot wkv)W_O.
\end{align}

In the Channel Mixing module, $R_c$ and $K_c$ are obtained similarly, while $V_c$ is computed as a linear projection of the activated $K_c$:

\begin{align}
    R_c &= Q\text{-}Shift_R(X)W_R \\
    K_c &= Q\text{-}Shift_K(X)W_K \\
    V_c &= SquaredReLU(K_c)W_V
\end{align}

The output $O_c$ is obtained by element-wise multiplication of $\sigma(R_c)$ and $V_c$, followed by a linear projection:

\begin{align}
    O_c = (\sigma(R_c) \odot V_c)W_O
\end{align}

The processed features from the Spatial Mixing and Channel Mixing modules are combined to yield the enhanced \underline{\textit{Semantic-BEV features}}, capturing both local and global representations for subsequent feature fusion.

\subsection{Occupancy Prediction Branch}
\label{sec:3b}

\noindent\textbf{\textit{Geo-RWKV Block}:} The occupancy prediction pathway (Fig. \ref{fig:block1}) is initialized with a $7 \times 7 \times 7$ convolutional layer, which is succeeded by a cascade of three Geo-RWKV blocks functioning as the encoder. These blocks exhibit a uniform architectural configuration, incorporating a residual connection that amalgamates both VRWKV and BEV projection modules. The VRWKV component executes spatial and channel mixing operations in accordance with the methodology established in the Sem-RWKV framework.

The residual structure commences by processing voxels $V \in \mathbb{R}^{1 \times L \times W \times H}$ derived from point cloud inputs, generating voxel representations that subsequently serve as the input $x$ for the BEV projection module. The three-dimensional dense features undergo alignment along the $z$-axis, followed by the application of two-dimensional convolutions for dimensional reduction, resulting in a set of dense \underline{\textit{Geometric-BEV features}} $\left \{  Geo^{bev,0}_f,Geo^{bev,1}_f,Geo^{bev,2}_f,Geo^{bev,3}_f  \right \}$. Through the exploitation of the VRWKV module's capacity for efficient long-range dependency modelling with linear computational complexity, the occupancy prediction pathway effectively processes and enhances the geometric information encoded within the voxel representation to generate refined \underline{\textit{Geometric-BEV features}}. These enhanced representations are subsequently utilized in the feature fusion process.

\begin{figure}[t]
  \centering
     \includegraphics[width=\linewidth]{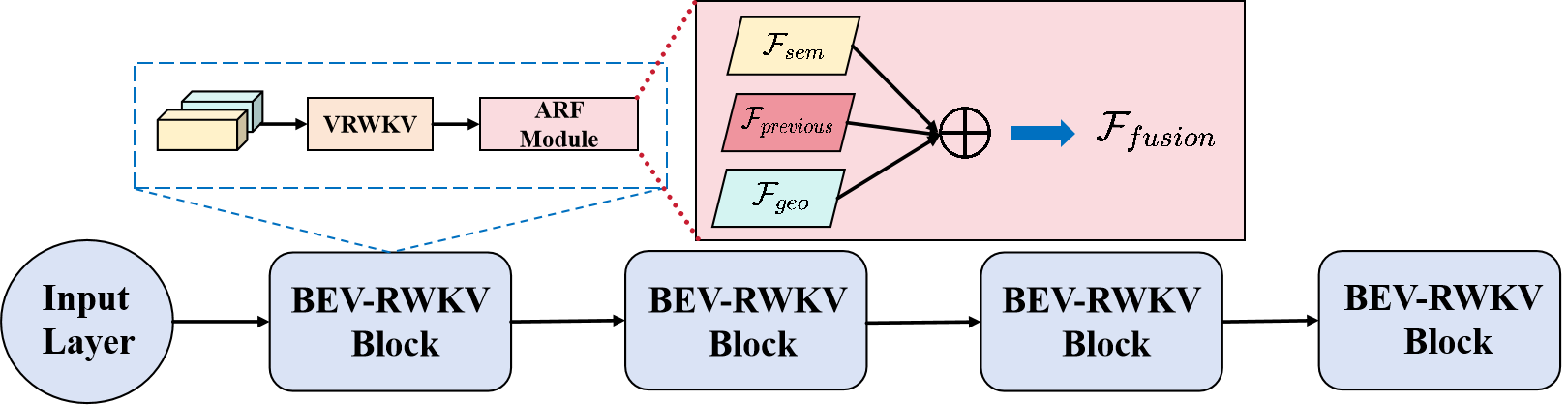}
   \caption{BEV feature fusion branch encoder structure. }
   \label{fig:block2}
\end{figure}

\begin{table*}[t]
		\footnotesize
		\setlength{\tabcolsep}{0.004\linewidth}
		\caption{Prediction results on SemanticKITTI test set. The C and L denote Camera and LiDAR, respectively.}
		
		\newcommand{\classfreq}[1]{{~\tiny(\semkitfreq{#1}\%)}}  %
		\centering
		\begin{tabular}{l|c|c| c c c c c c c c c c c c c c c c c c c | c}
			\toprule
			Method
                & \makecell[c]{Modality}
			& mIoU $\uparrow$
			& \rotatebox{90}{\textcolor{road}{$\blacksquare$} road}
			\rotatebox{90}{\ \ \ \classfreq{road}} 
			& \rotatebox{90}{\textcolor{sidewalk}{$\blacksquare$} sidewalk}
			\rotatebox{90}{\ \ \ \classfreq{sidewalk}}
			& \rotatebox{90}{\textcolor{parking}{$\blacksquare$} parking}
			\rotatebox{90}{\ \ \ \classfreq{parking}} 
			& \rotatebox{90}{\textcolor{other-ground}{$\blacksquare$} other-grnd}
			\rotatebox{90}{\ \ \ \classfreq{otherground}} 
			& \rotatebox{90}{\textcolor{building}{$\blacksquare$}  building}
			\rotatebox{90}{\ \ \ \classfreq{building}} 
			& \rotatebox{90}{\textcolor{car}{$\blacksquare$}  car}
			\rotatebox{90}{\ \ \ \classfreq{car}} 
			& \rotatebox{90}{\textcolor{truck}{$\blacksquare$}  truck}
			\rotatebox{90}{\ \ \ \classfreq{truck}} 
			& \rotatebox{90}{\textcolor{bicycle}{$\blacksquare$}  bicycle}
			\rotatebox{90}{\ \ \ \classfreq{bicycle}} 
			& \rotatebox{90}{\textcolor{motorcycle}{$\blacksquare$} motorcycle}
			\rotatebox{90}{\ \ \ \classfreq{motorcycle}} 
			& \rotatebox{90}{\textcolor{other-vehicle}{$\blacksquare$}  other-veh.}
			\rotatebox{90}{\ \ \  \classfreq{othervehicle}} 
			& \rotatebox{90}{\textcolor{vegetation}{$\blacksquare$} vegetation}
			\rotatebox{90}{\ \ \ \classfreq{vegetation}} 
			& \rotatebox{90}{\textcolor{trunk}{$\blacksquare$}  trunk}
			\rotatebox{90}{\ \ \ \classfreq{trunk}} 
			& \rotatebox{90}{\textcolor{terrain}{$\blacksquare$} terrain}
			\rotatebox{90}{\ \ \ \classfreq{terrain}} 
			& \rotatebox{90}{\textcolor{person}{$\blacksquare$}  person}
			\rotatebox{90}{\ \ \ \classfreq{person}} 
			& \rotatebox{90}{\textcolor{bicyclist}{$\blacksquare$}  bicyclist}
			\rotatebox{90}{\ \ \ \classfreq{bicyclist}} 
			& \rotatebox{90}{\textcolor{motorcyclist}{$\blacksquare$}  motorcyclist.}
			\rotatebox{90}{\ \ \ \classfreq{motorcyclist}} 
			& \rotatebox{90}{\textcolor{fence}{$\blacksquare$} fence}
			\rotatebox{90}{\ \ \ \classfreq{fence}} 
			& \rotatebox{90}{\textcolor{pole}{$\blacksquare$} pole}
			\rotatebox{90}{\ \ \ \classfreq{pole}} 
			& \rotatebox{90}{\textcolor{traffic-sign}{$\blacksquare$} traf.-sign}
			\rotatebox{90}{\ \ \ \classfreq{trafficsign}} 
                & FPS
			\\
			\midrule

    	MonoScene~\cite{cao2022monoscene}&C & 11.1 & 54.7 & 27.1 &24.8 & 5.7 & 14.4 & 18.8 & 3.3 & 0.5 & 0.7 & 4.4& 14.9 & 2.4 & 19.5 & 1.0 & 1.4 & 0.4 & 11.1 & 3.3 & 2.1 & 1.1 \\
			
            
            OccFormer~\cite{zhang2023occformer} & C   &12.3& 55.9& 30.3& 31.5& 6.5 &15.7 &21.6& 1.2& 1.5& 1.7 &3.2& 16.8& 3.9 &21.3& 2.2& 1.1& 0.2 &11.9 &3.8 &3.7  &1.8  \\
            
            VoxFormer~\cite{li2023voxformer} & C  & 13.4 & 54.1 & 26.9 & 25.1 & 7.3 & 23.5 & 21.7 &3.6 & 1.9 & 1.6 & 4.1 & 24.4 & 8.1 & 24.2 & 1.6 & 1.1 & 0.0 & 6.6 & 5.7 & 8.1 &1.5\\

            TPVFormer~\cite{huang2023tri} & C  & 11.3 & 55.1 & 27.2 & 27.4 & 6.5 & 14.8 & 19.2 & 3.7 & 1.0 & 0.5 & 2.3 & 13.9 & 2.6 & 20.4 & 1.1 & 2.4 & 0.3 & 11.0 & 2.9 & 1.5 & 1.0\\

            
            
            SSC-RS~\cite{mei2023ssc}&L    & 24.2 & 73.1 & 44.4 & 38.6 & \textbf{17.4} &\textbf{44.6} &36.4 &5.3 &10.1 &5.1& 11.2 & \textbf{44.1} &26.0 &41.9 &4.7& 2.4& 0.9& 30.8& 15.0& 7.2 & 16.7\\

            SCONet~\cite{wang2024agrnav}&L   & 17.6& 51.9 & 30.7 & 23.1 & 0.9 & 39.9 &29.1 &1.7 &0.8 &0.5& 4.8& 41.4 &27.5 &28.6 &0.8& 0.5& 0.1& 18.9& 21.4& 8.0 & 20.0\\
            
            JS3C-Net~\cite{yan2021sparse}&L  &23.8& 64.0& 39.0 & 34.2 & 14.7 & 39.4  &33.2 & 7.2 & 14.0 & \textbf{8.1}& \textbf{12.2}  &43.5&19.3 & 39.8 &\textbf{ 7.9}  &\textbf{5.2} & 0.0 & 30.1 & 17.9 &15.1  & 1.7 \\

            M-CONet~\cite{wang2023openoccupancy}&C\&L  & 20.4 & 60.6 & 36.1 &29.0  & 13.0 & 38.4 &33.8 & 4.7 &3.0 &2.2  & 5.9 & 41.5 &20.5 &35.1 & 0.8 & 2.3 & 0.6 & 26.0 &18.7 & 15.7  &1.4 \\
                   
            Co-Occ~\cite{pan2024co} &C\&L  &24.4  & 72.0  &43.5 & \textbf{42.5} & 10.2 &35.1  & \textbf{40.0}& 6.4 &4.4 &3.3 &8.8&41.2&\textbf{30.8}& 40.8 & 1.6 & 3.3 & 0.4 & \textbf{32.7}& \textbf{26.6}& \textbf{20.7}  & 1.1  \\
	      \midrule

             OccRWKV (Ours)&L  &\textbf{25.1}  & \textbf{73.5}  & \textbf{44.6} &40.2  & 16.8 & 42.8  & 35.5 & \textbf{7.3} & \textbf{14.1} & 7.9  &10.0  & 43.1  & 30.6  &\textbf{43.2} & 4.7 & 1.5 & \textbf{1.3} & 31.4 & 19.0 & 10.2 & \textbf{22.2}  \\
   
   \bottomrule
		\end{tabular}
		
		\vspace{1mm}
		
		\label{tab:kitti}
		
	\end{table*}

\subsection{BEV Feature Fusion Branch}
\label{sec:3c}

The BEV feature fusion branch adopts a U-Net architecture incorporating 2D convolutions and BEV-RWKV blocks (Fig. \ref{fig:block2}). Its encoder comprises an initial layer followed by four down-sampling stages, each integrated with a BEV-RWKV block. After processing the concatenated \underline{\textit{Semantic-BEV}} and \underline{\textit{Geometric-BEV}} features through the input layer and initial BEV-RWKV block, an ARF module \cite{mei2023ssc} fuses these multi-scale representations to capture both semantic context and geometric structure. The decoder utilizes up-sampling operations and skip connections for spatial detail reconstruction, ultimately generating a 3D semantic occupancy grid $\mathcal{O}\in \mathbb{R}^{((C_n+1)*L)*H*W}$, where $C_n$ denotes the class count.

\subsection{Loss Function}
\label{sec:3d}

Our loss function integrates 3 key elements. Specifically, the semantic loss component $L_{sems}$ aggregates the Lovasz loss \cite{berman2018lovasz} and the cross-entropy loss \cite{zhang2018generalized} at every stage within the semantic branch. For the occupancy branch, the training loss $L_{occ}$ is computed by summing the binary cross-entropy loss, $L_{binary\_cross}$, and the Lovasz loss at each respective stage, denoted by $i$. The BEV loss, $L_{bev}$, is defined as thrice the sum of the cross-entropy loss and the Lovasz loss. We train the entire network in an end-to-end manner. The overall objective function is:
\begin{equation}
L_{total} = L_{bev} + L_{sems} + L_{occ}
\end{equation}
subject to:
\begin{equation}
\small
\left\{
\begin{aligned}
& L_{sems} = \sum_{i=1}^{3}(L_{cross,i} + L_{lovasz,i}), \\
& L_{occ} = \sum_{i=1}^{3}(L_{binary\_cross,i} + L_{lovasz,i}), \\
& L_{bev} = 3\times (L_{cross} + L_{lovasz})
\end{aligned}
\right.
\end{equation}
where $L_{bev}$, $L_{sems}$, and $L_{occ}$ respectively represent the BEV loss, the semantic loss, and the occupancy loss.

\begin{figure}[t]
  \centering
     \includegraphics[width=0.96\linewidth]{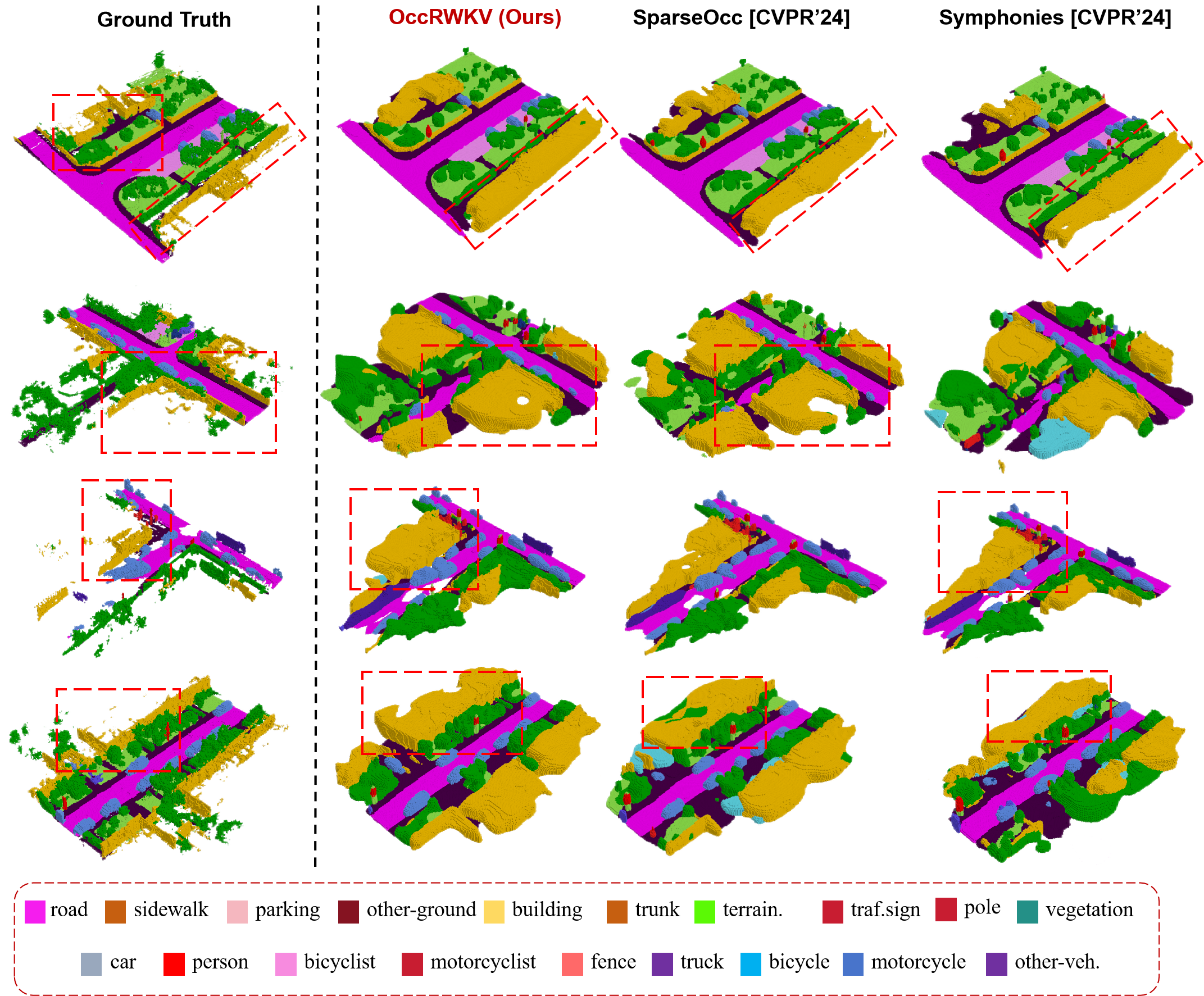}
   \caption{The qualitative comparisons results on the SemanticKITTI validation set.}
   \label{fig:exp1}
\end{figure}

\section{Experiments}
\label{sec:5}

\subsection{Experimental Setups}
\noindent\textbf{\textit{Dataset and Evaluation Metrics:}} OccRWKV trained on the SemanticKITTI dataset \cite{behley2019semantickitti} for semantic occupancy prediction using point clouds data, with ground truth represented in $[256, 256, 32]$ voxel grids. We evaluated the model using mean intersection over union (mIoU) for semantic accuracy and frames per second (FPS) for deployment feasibility on resource-constrained robots. The model was also tested for zero-shot reasoning on an aerial-ground robot\cite{wang2024agrnav}, demonstrating its potential to enhance navigation efficiency without prior environment-specific training.

\noindent\textbf{\textit{Implementation Details:}} OccRWKV was trained over 80 epochs with a batch size of 4 and an initial learning rate of 0.001 using the Adam optimizer \cite{kingma2014adam}, augmented by random flips along the $x-y$ axis. Post-training, the model was optimized with TensorRT and deployed on a Jetson Xavier NX for real-time occlusion perception in a robot's navigation system. The model's influence on navigation efficiency was appraised by conducting 10 trials across two varied scenes. For deployment specifics, please refer to the methodology outlined in \cite{wang2024agrnav}. 





\begin{table*}[t]
\centering
\caption{3D Occupancy Results on SemanticKITTI \cite{behley2019semantickitti} Validation Set}

\begin{tabular}{@{}lccccccc@{}}
\toprule
Method & IoU (\%) $\uparrow$ & mIoU (\%) $\uparrow$ & Precision (\%) $\uparrow$ & Recall (\%) $\uparrow$ & Parameters (M) $\downarrow$ & FLOPs (G) $\downarrow$ & Memory (GB) $\downarrow$ \\ 
\midrule
\multicolumn{8}{c}{\emph{MLP/CNN-based}} \\
\midrule
Monoscene \cite{cao2022monoscene} & 37.1 & 11.5 & 52.2 & 55.5 & 149.6 & 501.8 & 20.3 \\
NDC-Scene \cite{yao2023ndc}  & 37.2 & 12.7 & - & - & - & - & 20.1 \\
Symphonies \cite{jiang2023symphonize}  & 41.9 & 14.9 & 62.7 & 55.7 & 59.3 & 611.9 & 20.0 \\
SparseOcc \cite{tang2024sparseocc} & 36.5 & 13.1 & 49.8  & 58.1   & 203.6  & 393.0  & 13.0 \\
\midrule
\multicolumn{8}{c}{\emph{Transformer-based}} \\
\midrule
OccFormer \cite{zhang2023occformer} & 36.5 & 13.5 & 47.3 & 60.4 & 81.4 & 889.0 & 21.0\\
VoxFormer \cite{li2023voxformer} & 57.7 & 18.4 & 69.9 & \textbf{76.7} & 57.8 & - & 15.2 \\
TPVFormer \cite{huang2023tri}& 35.6 & 11.4 & - & -  & 48.8 & 946.0 & 20.0 \\
CGFormer \cite{yu2024context}& 45.9 & 16.9 & 62.8 & 63.2 & 122.4 & \textbf{314.5} & 19.3 \\
\midrule
\multicolumn{8}{c}{\emph{RWKV-based (Ours)}} \\
\midrule
\textbf{OccRWKV} & \textbf{58.8}  & \textbf{25.0}  & \textbf{78.1}  & 70.4 & \textbf{37.9} & 397.6 & \textbf{7.1} \\
\bottomrule
\end{tabular}%

\label{tab:3d_occupancy_results}
\end{table*}

\subsection{OccRWKV Comparison against the state-of-the-art.}
\label{sec:B1}
\noindent\textbf{\textit{Quantitative Results:}}  OccRWKV sets a new benchmark on the SemanticKITTI hidden test dataset (Table~\ref{tab:kitti}), with a 25.1\% mIoU, surpassing the leading camera-based algorithm, LowRankOcc \cite{zhao2024lowrankocc}, by 84.6\% and the foremost LiDAR-based technique, SSC-RS \cite{mei2023ssc}, by 3.7\%. Regarding processing efficiency, OccRWKV achieves an impressive FPS of 22.2, more than 22 times faster than Co-Occ \cite{pan2024co}. This efficiency, combined with superior accuracy, underscores the advantages of OccRWKV over fusion-based methods, highlighting its robustness and the benefits of a LiDAR-centric approach for real-time navigational tasks in robotics. 

We also have conducted comparative evaluations using established CNN-based and Transformer-based methods. The results, as presented in Table~\ref{tab:3d_occupancy_results}, indicate that OccRWKV achieves superior performance on the SemanticKITTI validation set, with an IoU of 58.8 and a mIoU of 25.0, surpassing the benchmark figures of the most notable studies within these two categories. Meanwhile, OccRWKV distinguishes itself with a parameter size of just 37.9 MB, which is 81.36\% smaller than the cutting-edge SparseOcc \cite{tang2024sparseocc}, making it significantly more efficient for deployment. Regarding computational resources, it requires only 7.1 GB of GPU memory, further emphasizing its practicality for real-world applications.

\noindent\textbf{\textit{Qualitative Results:}} Fig.~\ref{fig:exp1} showcases the 3D semantic occupancy predictions from OccRWKV for various intricate environments within the SemanticKITTI validation set. Notably, OccRWKV more effectively reconstructs expansive, flat road surfaces and accurately captures intricate features such as distant vegetation and moving vehicles. The success of OccRWKV can be attributed to the innovative RWKV-based tri-branch network architecture, which facilitates the generation of precise, scene-level representations efficiently.  Such capability proves highly beneficial for robotic navigation tasks, enabling proactive discernment of obstacle layouts in obscured areas and the formulation of comprehensive local maps.

\begin{table}[t]
\caption{\small Ablation Study on SemanticKITTI  Validation Set.}
\small
\centering
\footnotesize
\resizebox{\columnwidth}{!}{%
\begin{tabular}{lccccc}
\toprule
Method & IoU $\uparrow$ & mIoU $\uparrow$ & Prec. & Recall & F1 \\
\midrule
OccRWKV & 58.8 & 25.0 & 78.0 & 70.4 & 74.0 \\
w/o Geo-RWKV Block & 58.2 & 24.1 & 77.5 & 69.9 & 73.8 \\
w/o Sem-RWKV Block & 57.6 & 23.4 & 77.1 & 69.2 & 73.0 \\
w/o BEV-RWKV Block & 57.9 & 23.9 & 76.7 & 68.9 & 72.4 \\
\bottomrule
\end{tabular}%
}
\label{tab:ablation}
\end{table}

\begin{figure}[t]
  \centering
     \includegraphics[width=0.95\linewidth]{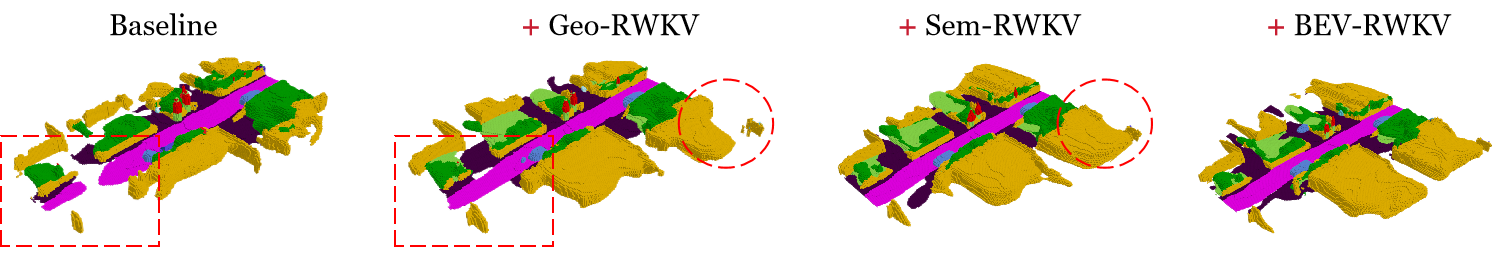}
   \caption{\ The visualization of ablation study on the impact of different components in the SemanticKITTI validation set.}
   \label{fig:ablation}
\end{figure}

\noindent\textbf{\textit{Ablation Study:}} Ablation studies on the SemanticKITTI set (Table~\ref{tab:ablation}) reveal the Sem-RWKV, Geo-RWKV, and BEV-RWKV modules' vital roles in our network. Sem-RWKV's removal notably decreases mIoU by 6.4\%, affirming its importance in detailed semantic segmentation. As Fig.~\ref{fig:ablation} shows, combining Sem-RWKV and Geo-RWKV enhances scene prediction accuracy by capturing long-range dependencies. The BEV-RWKV's impact on metrics is minor, serving mainly to reduce computational load during feature fusion.

\subsection{Impact of OccRWKV on real-world navigation performance.}
\label{sec:D1}

\begin{figure}[t]
  \centering
     \includegraphics[width=\linewidth]{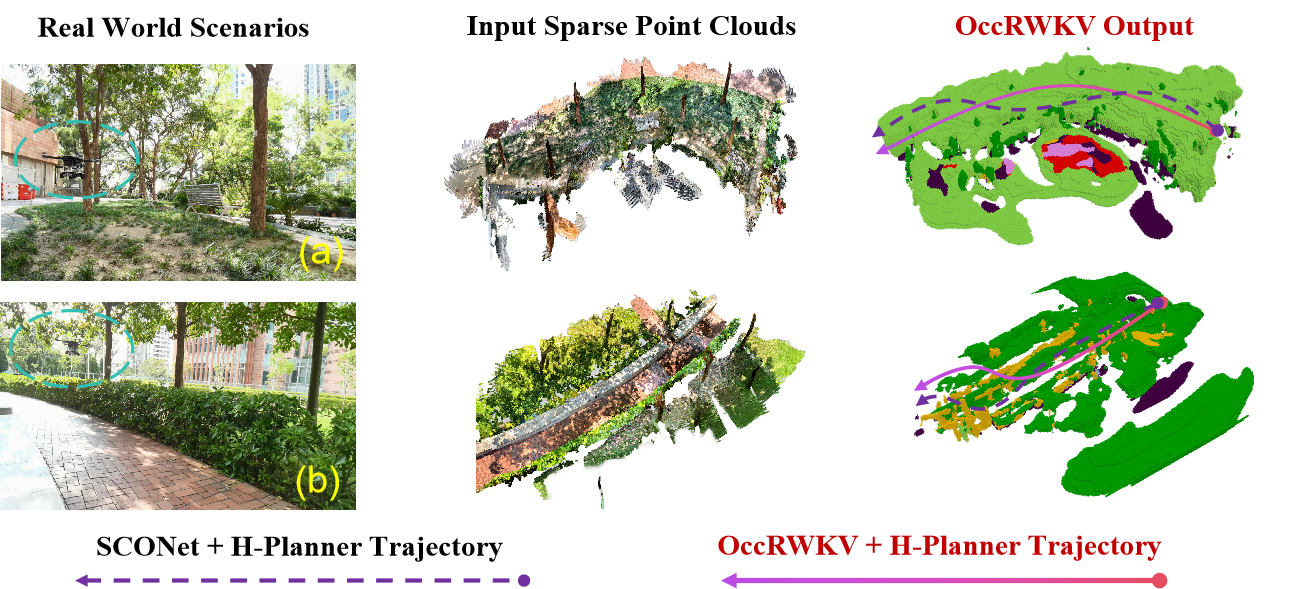}
   \caption{OccRWKV is deployed offline on a robot for zero-shot semantic occupancy prediction.  }
   \label{fig:real-world}
\end{figure}

We integrated the OccRWKV model, previously trained on the SemanticKITTI dataset, into an aerial-ground robot's navigation system to serve as its perception network (i.e., replace SCONet from AGRNav\cite{wang2024agrnav}). Following the objectives outlined in \cite{wang2024agrnav}, the model preemptively predicts the distribution of obstacles in obscured areas to produce a complete local map, facilitating faster robot traversal.  Experiments across 2 occlusion environments (Table~\ref{tab:real}) showed the average movement time without a perception network was 23.92 seconds. With the inclusion of the perception network from \cite{wang2024agrnav}, this time was reduced to 16.54 seconds. The application of OccRWKV further improved results, cutting movement time down to 13.79 seconds and decreasing energy consumption. This efficiency gain is attributed to the detailed local maps generated by OccRWKV, thereby curtailing flight paths. Moreover, as depicted in Fig.~\ref{fig:real-world}, OccRWKV exhibits strong zero-shot 3D semantic occupancy prediction, yielding dense predictions from sparse point clouds and precisely identifying semantic elements like vegetation and roads.

\begin{table}[t]
\centering
\caption{\small Impact of OccRWKV on navigation efficiency. }
\footnotesize
\label{tab:real}
\begin{tabular}{@{}cccc@{}}
\toprule
Perception & Planner & Move. Time (s) & Ener. Con (J)\\
\midrule
- & H-Planner \cite{wang2024agrnav}    & 23.92   &  15362.79 \\
SCONet \cite{wang2024agrnav} & H-Planner \cite{wang2024agrnav}    & 16.54 & 12380.33 \\
\bottomrule
 OccRWKV & H-Planner \cite{wang2024agrnav}  & \textbf{13.79}   & \textbf{11625.98}  \\
\bottomrule
\end{tabular}
\end{table}

\section{CONCLUSIONS}

In conclusion, OccRWKV, our novel network, successfully addresses the challenge of balancing performance and efficiency in 3D semantic occupancy prediction. It delivers state-of-the-art accuracy with a mIoU of 25.1 on the SemanticKITTI benchmark and maintains efficient real-time performance at 22.2 FPS. The network's scalability makes it a robust solution for practical applications in robot navigation and autonomous driving. Field deployments confirm OccRWKV's effectiveness in real-world settings, validating its suitability for future integration in complex environments.










\newpage

\bibliographystyle{IEEEtran}
\balance
\bibliography{root}

\end{document}